\documentclass{article}

\usepackage[preprint,nonatbib]{nips_2018}
\usepackage[utf8]{inputenc} 
\usepackage[T1]{fontenc}    
\usepackage{hyperref}       
\usepackage{url}            
\usepackage{booktabs}       
\usepackage{amsfonts}       
\usepackage{nicefrac}       
\usepackage{microtype}      

\usepackage{cite}
\usepackage{amsmath}
\usepackage{enumitem}
\usepackage{graphicx}
\usepackage[all]{hypcap}
\graphicspath{{figs/}}
\newcommand{\figref}[1]{\figurename~\ref{#1}}

\newcommand{\secref}[1]{Section~\ref{#1}}

\title{Training Generative Adversarial Networks with Binary Neurons by End-to-end Backpropagation}

\author{Hao-Wen Dong, Yi-Hsuan Yang\\%
Research Center for IT Innovation, Academia Sinica, Taipei, Taiwan\\%
\texttt{\{salu133445,yang\}@citi.sinica.edu.tw}%
}

\begin{document}

\maketitle

\begin{abstract}
We propose the BinaryGAN, a novel generative adversarial network (GAN) that uses binary neurons at the output layer of the generator. We employ the sigmoid-adjusted straight-through estimators to estimate the gradients for the binary neurons and train the whole network by end-to-end backpropogation. The proposed model is able to directly generate binary-valued predictions at test time. We implement such a model to generate binarized MNIST digits and experimentally compare the performance for different types of binary neurons, GAN objectives and network architectures. Although the results are still preliminary, we show that it is possible to train a GAN that has binary neurons and that the use of gradient estimators can be a promising direction for modeling discrete distributions with GANs. For reproducibility, the source code is available at \url{https://github.com/salu133445/binarygan}.
\end{abstract}

\section{Introduction}

Generative adversarial networks (GAN)~\cite{gan} have enjoyed great success in modeling continuous distributions. However, applying GANs to discrete data have been shown nontrivial in that it is difficult to optimize the model distribution toward the target data distribution in a high-dimensional discrete space. Approaches adopted in the literature to utilize GANs on discrete data can roughly be divided into three directions.

One direction is to replace the target discrete outputs with continuous relaxations. Kusner \textit{et al.}~\cite{gumbel_softmax} proposed to use the continuous Gumbel-Softmax distribution to approximate a categorical distribution and generate sequences of discrete elements using one-hot encoding. Using the Wasserstein GANs~\cite{wgan}, Gulrajani \textit{et al.}~\cite{wgan-gp} and Subramanian \textit{et al.}~\cite{adv_nlp} have developed in parallel models that can handle discrete data by simply passing the continuous, probabilistic outputs (i.e., softmax relaxation) of the generator to the discriminator.

The second direction is to view the generator as an agent in reinforcement learning (RL) and introduce RL-based training strategies. Yu \textit{et al.}~\cite{seqgan} considered the generator as a stochastic parametrized policy and trained the generator via policy gradient and Monte Carlo search. Hjelm \textit{et al.}~\cite{bgan} used estimated difference measure from the discriminator to compute importance weights for generated samples, which provides a policy gradient for training the generator. More examples can be seen in natural language processing (NLP), including dialogue generation~\cite{adv_dialogue} and machine translation~\cite{adv_nmt}.

The third direction is to introduce gradient estimators to estimate the gradients for the nondifferentiable discretization operations for the generator. Since the discretization operations are used in the forward pass, the generator is able to provide discrete predictions to the discriminator during the training and directly generate discrete predictions at test time without any further post-processing. Moreover, it can support conditional computation graphs~\cite{bengio,hmrnn} that allow the system to make discrete decisions for more advanced designs. However, to the best of our knowledge and as pointed out by~\cite{bgan}, none of these has yet been shown to work with GANs.

\begin{figure}
\centering
\includegraphics[width=.8\linewidth]{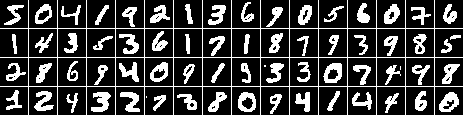}
\caption{Sample binarized MNIST digits seen in our training data.}
\label{fig:train}
\end{figure}

In our previous work on generating binary-valued music pianorolls~\cite{bmusegan}, we proposed to append to the generator a refiner network that learns to binarize the generator's outputs to binary ones. However, the whole network was trained in a two-stage setting. It remains unclear whether and how we can train a GAN that has binary neurons in an end-to-end manner. We intend to study this issue in this paper and consider here the generation of binarized MNIST handwritten digits (see \figref{fig:train}) as a case study, assuming that people are more familiar with the MNIST digits than the pianorolls.

In this paper, we employ either stochastic or deterministic binary neurons at the output layer of the generator. In order to train the whole network by end-to-end backpropagation, we use the sigmoid-adjusted straight-through estimators~\cite{stestimator,bengio,r2rt} to estimate the gradients for the binary neurons. We experimentally compare the performance of the proposed model, which we dub BinaryGAN, using different types of binary neurons, GAN objectives and network architectures.

\section{Backgrounds}

\subsection{Generative Adversarial Networks}
\label{sec:gan}

A generative adversarial network (GAN)~\cite{gan} is a generative model that can learn a data distribution in an unsupervised manner. It contains two components: a \textit{generator} $G$ and a \textit{discriminator} $D$. The generator takes as input a random vector $\mathbf{z}$ sampled from a prior distribution $p_\mathbf{z}$ and outputs a fake sample. The discriminator take as input either a real sample drawn from the data distribution $p_d$ or a fake sample generated by and outputs a scalar representing the genuineness of that sample.

The training is formulated as a two-player game: the discriminator aims to tell the fake data from the real ones, while the generator aims to fool the discriminator. Note that in this paper we refer to GAN as its non-saturating version, which can provide stronger gradients in the early stage of the training as suggested by~\cite{gan}. The objectives for the generator and the discriminator are
\begin{align}
&\textbf{(GAN)} \quad \min_{G} -\mathbf{E}_{\mathbf{z}\sim p_\mathbf{z}}[\log D(G(\mathbf{z})),\\
&\textbf{(GAN)} \quad \max_{D} \mathbf{E}_{\mathbf{x}\sim p_d}[\log D(\mathbf{x})] + \mathbf{E}_{\mathbf{z}\sim p_\mathbf{z}}[\log(1 - D(G(\mathbf{z})))],
\end{align}
Another form called WGAN~\cite{wgan} was later proposed with the intuition to estimate the Wasserstein distance between the real and the model distributions by a deep neural network and use it as a critic for the generator. It can be formulated as
\begin{equation}
\textbf{(WGAN)} \quad \min_{G} \max_{D} \mathbf{E}_{\mathbf{x}\sim p_d}[D(\mathbf{x})] - \mathbf{E}_{\mathbf{z}\sim p_\mathbf{z}}[D(G(\mathbf{z}))]\,,
\end{equation}
where the discriminator $D$ must be Lipschitz continuous.

In~\cite{wgan}, the Lipschitz constraint on the discriminator is imposed by \textit{weight clipping}. However, this can lead to undesired behaviors as discussed in~\cite{wgan-gp}. A \textit{gradient penalty} (GP) term that punishes the discriminator when it violates the Lipschitz constraint is then proposed in~\cite{wgan-gp}. The objectives become
\begin{align}
&\textbf{(WGAN-GP)} \quad \min_{G}\mathbf{E}_{\mathbf{x}\sim p_d}[D(\mathbf{x})] - \mathbf{E}_{\mathbf{z}\sim p_\mathbf{z}}[D(G(\mathbf{z}))]\,,\\
&\textbf{(WGAN-GP)} \quad \max_{D} \mathbf{E}_{\mathbf{x}\sim p_d}[D(\mathbf{x})] - \mathbf{E}_{\mathbf{z}\sim p_\mathbf{z}}[D(G(\mathbf{z}))] - \mathbf{E}_{\hat{\mathbf{x}}\sim p_{\hat{\mathbf{x}}}} [(\nabla_{\hat{\mathbf{x}}}\|\hat{\mathbf{x}}\|-1)^2]\,,
\end{align}
where $p_{\hat{\mathbf{x}}}$ is defined sampling uniformly along straight lines between pairs of points sampled from $p_d$ and $p_g$, the model distribution.

\subsection{Deterministic and Stochastic Binary Neurons}
\label{sec:bn}

Binary neurons are neurons that output binary-valued predictions. In this work, we consider two types of them. A \textit{deterministic binary neuron} (DBN) acts like a neuron with the hard thresholding function as its activation function. The output of a DBN for a real-valued input $x$ is defined as
\begin{equation}
DBN(x) = u(\sigma(x)-0.5)\,,
\label{eq:dbn}
\end{equation}
where $u(\cdot)$ denotes the unit step function and $\sigma(\cdot)$ is the sigmoid function.

A \textit{stochastic binary neuron} (SBN) acts like a neuron using Bernoulli sampling as its activation function and binarizes a real-valued input $x$ according to probability, defined as
\begin{equation}
SBN(x) = u(\sigma(x) - v),\; v\sim U[0, 1]\,,
\label{eq:sbn}
\end{equation}
where $U[0, 1]$ denotes a uniform distribution. Note that in this work we refer to $\sigma(x)$ in \eqref{eq:dbn} and \eqref{eq:sbn} as the \textit{preactivated outputs}.

\subsection{Sigmoid-adjusted Straight-through Estimators}

Backpropgating through binary neurons, however, is intractable. The reason is that, for a DBN, the threshold function is nondifferentiable and that, for an SBN, it requires the computation of the expected gradient averaged on all possible combinations of values taken by the binary neurons, where the number of such combinations is exponential to the total number of binary neurons.

The \textit{straight-through estimator} was first proposed as a regularizer in~\cite{stestimator}. It simply ignores the gradient of a binary neuron and treats it as an identity function in the backward pass. A variant called \textit{sigmoid-adjusted straight-through estimator}~\cite{bengio} replaces the gradient of a binary neuron with the gradient of the sigmoid function instead. The latter is found to achieve better performance in a classification task presented in~\cite{r2rt}. Hence, when training networks with binary neurons, we resort to the sigmoid-adjusted straight-through estimators to provide the gradients for the binary neurons.

\section{BinaryGAN}

\begin{figure}
\centering
\includegraphics[width=.8\linewidth]{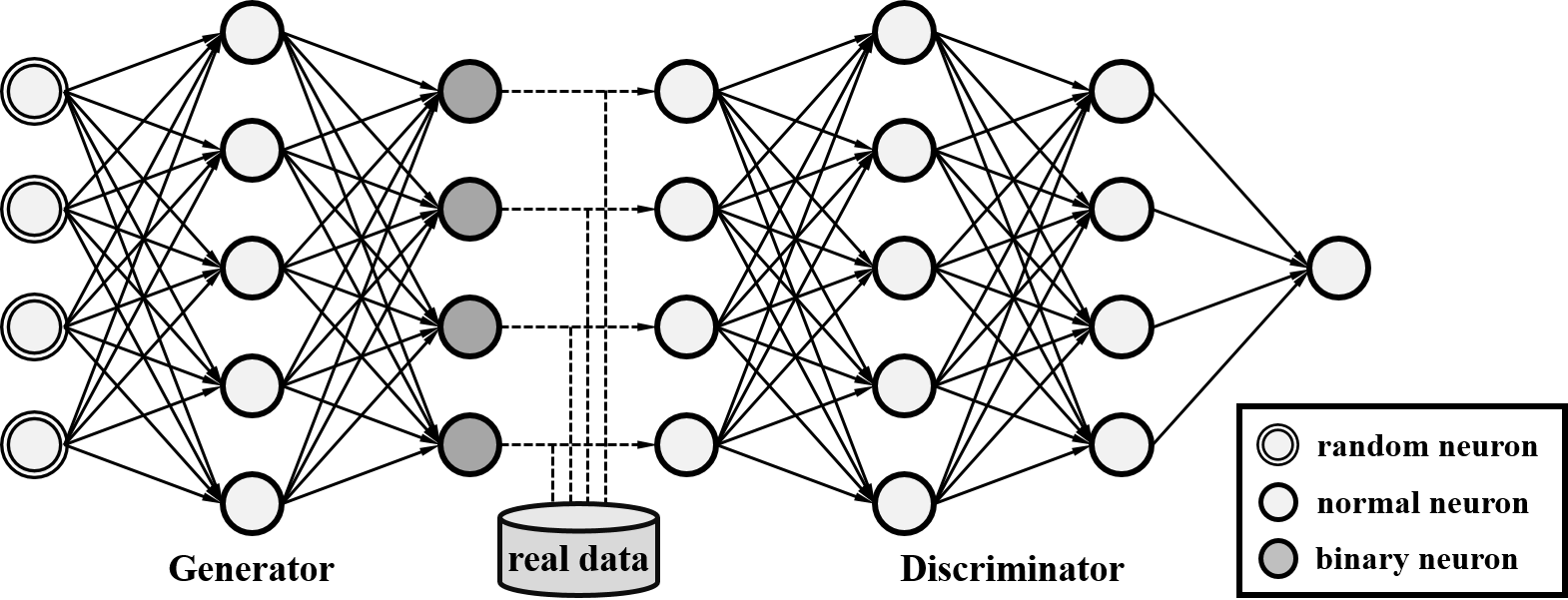}
\caption{System diagram of the proposed model implemented by MLPs. Note that binary neurons are only used at the output layer of the generator.}
\label{fig:system}
\end{figure}

We propose the BinaryGAN, a model that can generate binary-valued predictions without any further post-processing and can be trained by end-to-end backpropagation. The proposed model consists of a \textit{generator} $G$ and a \textit{discriminator} $D$. The generator takes as input a random vector $\mathbf{z}$ drawn from a prior distribution $p_\mathbf{z}$ and generate a fake sample $G(\mathbf{z})$. The discriminator takes as input either a real sample drawn from the data distribution or a fake sample generated by the generator and outputs a scalar indicating the genuineness of that sample.

In order to handle binary data, we propose to use binary neurons, either deterministic or stochastic ones, at the \textit{output layer} (i.e., the final layer) of the generator. Hence, the model space (i.e., the output space of the generator) is $2^N$, where $N$ is the number of visible binary neurons at the output layer. We employ the sigmoid-adjusted straight-through estimators to estimate the gradients for binary neurons and train the whole network by end-to-end backpropagation. \figref{fig:system} shows the system diagram for the proposed model implemented by multilayer perceptrons (MLPs).

\begin{table}
\centering
\begin{minipage}{.43\linewidth}
\centering
\begin{tabular}{lcc}
  \toprule
  Layer &Number of nodes &Activation\\
  \midrule
  \textit{dense} &$1024$ &ReLU\\
  \textit{dense} &$784$  &sigmoid\\
  \bottomrule
\end{tabular}\\[1ex]
\caption{Network architecture of the generator for the MLP model.}
\label{tab:net_mlp_g}
\end{minipage}
\hspace{6ex}
\begin{minipage}{.45\linewidth}
\centering
\begin{tabular}{lcc}
  \toprule
  Layer &Number of nodes &Activation\\
  \midrule
  \textit{dense} &$512$ &LeakyReLU\\
  \textit{dense} &$256$ &LeakyReLU\\
  \textit{dense} &$1$   &sigmoid\\
  \bottomrule
\end{tabular}\\[1ex]
\caption{Network architecture of the discriminator for the MLP model.}
\label{tab:net_mlp_d}
\end{minipage}
\end{table}

\section{Experiments and Results}

\subsection{Training Data---Binarized MNIST Database}

In this work, we use the binarized version of the MNIST handwritten digit database~\cite{mnist}. Specifically, we convert pixels with nonzero intensities to ones and pixels with zero intensities to zeros. \figref{fig:train} shows some sample binarized MNIST digits seen in our training data.

\subsection{Implementation Details}

\begin{itemize}
\item Both the generator and the discriminator are implemented as multilayer perceptrons (MLPs) (see Tables \ref{tab:net_mlp_g} and \ref{tab:net_mlp_d} for the network architectures). Note that other network architectures will be compared in \secref{sec:exp_net}.
\item The batch size for all the experiments is set to $64$.
\item As suggested by~\cite{wgan-gp}, we apply batch normalization~\cite{batchnorm} only to the generator and omit batch normalization in the discriminator.
\item We train the proposed model with the WGAN-GP objective. Note that other GAN objectives will be compared in \secref{sec:exp_gan}.
\item We use the Adam optimizer~\cite{adam}, as suggested by~\cite{wgan-gp}, with hyperparameters $\alpha = 0.0001, \beta_1 = 0.5$ and $\beta_2 = 0.9$.
\item We apply the \textit{slope annealing trick}~\cite{hmrnn}. Specifically, we multiply the slopes of the sigmoid functions in the sigmoid-adjusted straight-through estimators by $1.1$ after each epoch.
\end{itemize}

Our implementation of binary neurons are mostly based on the code provided in the blog post---``Binary Stochastic Neurons in Tensorflow''---on the R2RT blog~\cite{r2rt}.

\subsection{Experiment I---Comparison of the proposed model using deterministic binary neurons and stochastic binary neurons}

\begin{figure}
\centering
\begin{minipage}{.49\linewidth}
\centering
\includegraphics[width=.75\linewidth]{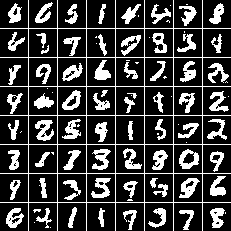}\\
(a) proposed model with DBNs\\[1ex]
\includegraphics[width=.75\linewidth]{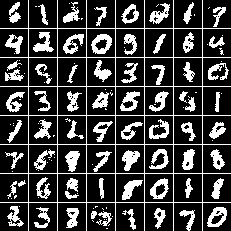}\\
(c) proposed model with SBNs
\end{minipage}
\hfill
\begin{minipage}{.49\linewidth}
\centering
\includegraphics[width=.75\linewidth]{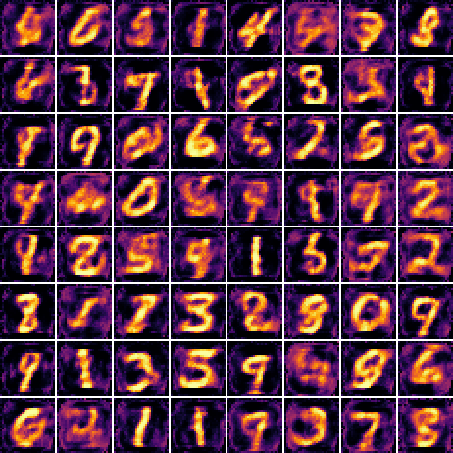}\\
(b) proposed model with DBNs (preactivated)\\[1ex]
\includegraphics[width=.75\linewidth]{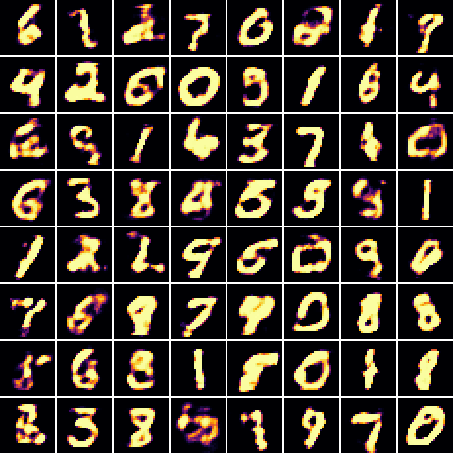}\\
(d) proposed model with SBNs (preactivated)
\end{minipage}\\[1ex]
\includegraphics[width=.4\linewidth,trim={0 15pt 0 0}]{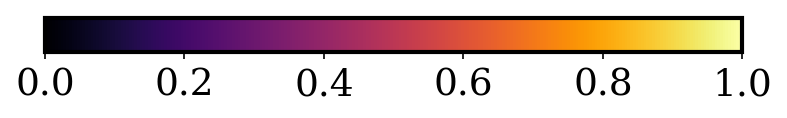}
\caption{Sample generated digits and preactivated outputs (i.e., the real-valued, intermediate values right before the binarization operation; see \secref{sec:bn}) for the proposed model implemented by MLPs and trained with the WGAN-GP objective.}
\label{fig:mlp}
\end{figure}

\begin{figure}
\centering
\includegraphics[width=.8\linewidth,trim={0 12pt 0 0}]{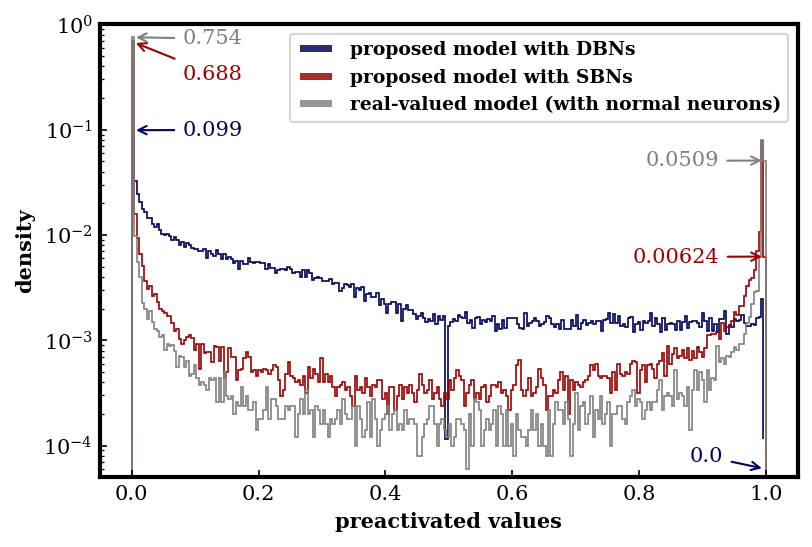}
\caption{Histograms of the preactivated outputs for the proposed model and the probabilistic predictions for the real-valued model. The two models are both implemented by MLPs and trained with the WGAN-GP objective.}
\label{fig:histogram}
\end{figure}

In the first experiment, we compare the performance of using \textbf{deterministic binary neurons} (\textbf{DBNs}) and \textbf{stochastic binary neurons} (\textbf{SBNs}) in the proposed model. We show in Figures \ref{fig:mlp}(a) and \ref{fig:mlp}(c) some sample generated digits for the two models. We can see that the proposed model with DBNs and SBNs can achieve similar qualities. However, from Figures \ref{fig:mlp}(b) and \ref{fig:mlp}(d) we can see that the \textit{preactivated outputs} (i.e., the real-valued, intermediate values right before the binarization operation; see \secref{sec:bn}) for the two models show distinct characteristics. In order to see how DBNs and SBNs work differently, we compute the histograms of their preactivated outputs, as shown in \figref{fig:histogram}.

We can see from \figref{fig:histogram} that the proposed model with DBNs outputs more preactivated values in the middle of zero and one, which results in a flatter histogram. We attribute this phenomenon to the fact that the output of a DBN is less sensitive to the absolute value for it depends only on whether the preactivated value is greater than the threshold. Moreover, we observe a notch around $0.5$, the threshold value we use in our implementation. It seems that DBNs tend to avoid producing preactivated values around the decision boundary (i.e., the threshold).

In contrast, the proposed model with SBNs outputs more preactivated values close to zero and one, which we attribute to the fact that the output of an SBN is more sensitive to the absolute value for it relies on Bernoulli sampling (e.g., an SBN may fire even with a tiny preactivated value). As a result, in order to avoid false triggering, it seems that an SBN tends to produce a preactivated value closer to zero and one.

\subsection{Experiment II---Comparison of the proposed model and the real-valued model}

\begin{figure}
\centering
\hspace{1.5cm}\includegraphics[width=0.35\linewidth]{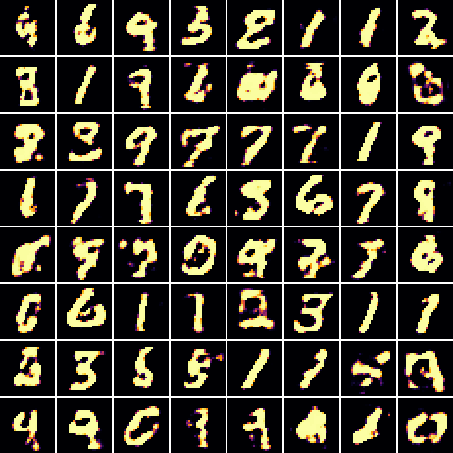}\hspace{2ex}\includegraphics[scale=.55]{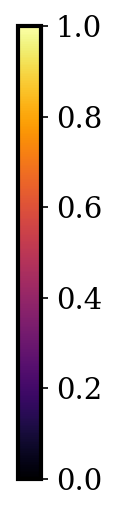}\\
(a) raw predictions\\[1ex]
\begin{minipage}{.49\linewidth}
\centering
\includegraphics[width=.75\linewidth]{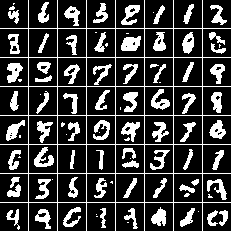}\\
(b) hard thresholding
\end{minipage}
\hfill
\begin{minipage}{.49\linewidth}
\centering
\includegraphics[width=.75\linewidth]{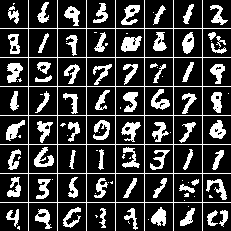}\\
(c) Bernoulli sampling
\end{minipage}
\caption{Sample raw predictions and binarized results for the real-valued model implemented by MLPs and trained with the WGAN-GP objective. (a) shows the raw, probabilistic outputs. (b) and (c) show the binarized results of applying a threshold of $0.5$ and Bernoulli sampling, respectively, to the raw predictions shown in (a).}
\label{fig:mlp_real}
\end{figure}

In the second experiment, we compare the proposed model with a variant that uses normal neurons at the output layer (with sigmoid functions as the activation functions).\footnote{This is how we train the MuseGAN model in~\cite{musegan}. After the training, we binarize the real-valued predictions with a threshold of $0.5$ to obtain the final binary-valued results.} We refer to this model as the \textbf{real-valued model}. \figref{fig:mlp_real}(a) shows some sample raw, probabilistic predictions of this model. Figures \ref{fig:mlp_real}(b) and \ref{fig:mlp_real}(c) show the final binarized results using two common post-processing strategies: \textit{hard thresholding} and \textit{Bernoulli sampling}, respectively. 

We also show in \figref{fig:histogram} the histogram of its probabilistic predictions. We can see that the histogram of this real-valued model is more U-shaped than that of the proposed model with SBNs. Moreover, there is no notch in the middle of the curve as compared to the proposed model with DBNs. From here we can see how different binarization strategies can shape the characteristics of the preactivated outputs of binary neurons. This also emphasizes the importance of including the binarization operations in the training so that the binarization operations themselves can also be optimized.

\subsection{Experiment III---Comparison of the proposed model trained with the GAN, WGAN and WGAN-GP objectives}
\label{sec:exp_gan}

In the third experiment, we compare the proposed model trained by the \textbf{WGAN-GP} objective with that trained by the \textbf{GAN} objective and by the \textbf{WGAN} objective (using weight clipping). Implementation details are summarized as follows.

\begin{itemize}
\item For the GAN model, we apply batch normalization to the generator and the discriminator.
\item For the WGAN model, we find it works better to apply batch normalization to the generator while omitting it for the discriminator.
\item For the GAN model, we employ the Adam optimizer~\cite{adam} with hyperparameters $\alpha = 0.0001, \beta_1 = 0.5$ and $\beta_2 = 0.9$.
\item For the WGAN model, we use the RMSProp optimizer~\cite{rmsprop}, as suggested by~\cite{wgan}, with a learning rate of $0.0001$ and a decay rate of $0.9$.
\end{itemize}

\begin{figure}
\centering
\begin{minipage}{.49\linewidth}
\centering
\includegraphics[width=.75\linewidth]{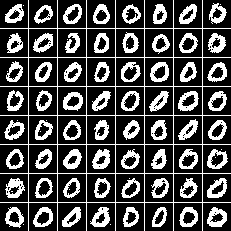}\\
(a) GAN model (with DBNs)\\[1ex]
\includegraphics[width=.75\linewidth]{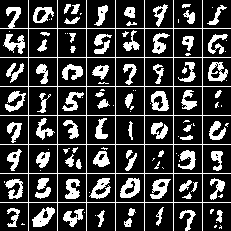}\\
(c) WGAN model (with DBNs)
\end{minipage}
\hfill
\begin{minipage}{.49\linewidth}
\centering
\includegraphics[width=.75\linewidth]{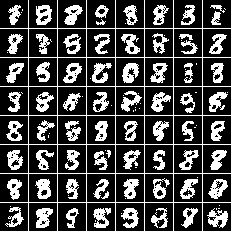}\\
(b) GAN model (with SBNs)\\[1ex]
\includegraphics[width=.75\linewidth]{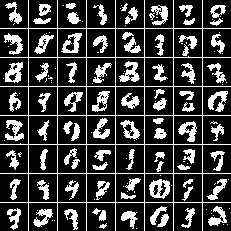}\\
(d) WGAN model (with SBNs)
\end{minipage}
\caption{Sample generated digits for the GAN and WGAN models, both implemented by MLPs.}
\label{fig:exp_gan}
\end{figure}

As can be seen from~\figref{fig:exp_gan}, the WGAN model is able to generate digits with similar qualities as the WGAN-GP model does, while the GAN model suffers from the so-called \textit{mode collapse} issue.

\subsection{Experiment IV---Comparison of the proposed model using multilayer perceptrons and convolutional neural networks}
\label{sec:exp_net}

In the last experiment, we compare the performance of using \textbf{multilayer perceptrons} (\textbf{MLPs}) and \textbf{convolutional neural networks} (\textbf{CNNs}). For the CNN model, we implement both the generator and the discriminator as deep CNNs (see Tables \ref{tab:net_cnn_g} and \ref{tab:net_cnn_d} for the network architectures). Note that the number of trainable parameters for the MLP and CNN models are 0.53M and 1.4M, respectively.

\begin{table}[t]
\centering
\begin{tabular}{lccccc}
  \toprule
  Layer &Number of filters &Kernel &Strides &Activation\\
  \midrule
  \textit{transconv} &$128$ &$2\times2$ &$(1, 1)$ &ReLU\\
  \textit{transconv} &$64$  &$4\times4$ &$(2, 2)$ &ReLU\\
  \textit{transconv} &$32$  &$3\times3$ &$(2, 2)$ &ReLU\\
  \textit{transconv} &$1$   &$4\times4$ &$(2, 2)$ &sigmoid\\
  \bottomrule
\end{tabular}\\[1ex]
\caption{Network architecture of the generator for the CNN model.}
\label{tab:net_cnn_g}
\end{table}

\begin{table}[t]
\centering
\begin{tabular}{lccccc}
  \toprule
  Layer &Number of filters &Kernel &Strides &Activation\\
  \midrule
  \textit{conv}    &$32$ &$3\times3$ &$(1, 1)$ &LeakyReLU\\
  \textit{maxpool} &     &$2\times2$ &$(2, 2)$ &\\
  \textit{conv}    &$64$ &$3\times3$ &$(1, 1)$ &LeakyReLU\\
  \textit{maxpool} &     &$2\times2$ &$(2, 2)$ &\\
  \textit{flatten}\\
  \textit{dense}   &$128$ &&&LeakyReLU\\
  \textit{dense}   &$1$   &&&sigmoid\textsuperscript{*}\\
  \bottomrule
\end{tabular}\\[.5ex]
{\footnotesize \textsuperscript{*}No activation for the WGAN and WGAN-GP models.}\\[1ex]
\caption{Network architecture of the discriminator for the CNN model.}
\label{tab:net_cnn_d}
\end{table}

\begin{figure}
\centering
\begin{minipage}{.49\linewidth}
\centering
\includegraphics[width=0.75\linewidth]{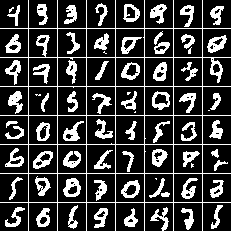}\\
(a)proposed (with DBNs)\\[1ex]
\includegraphics[width=0.75\linewidth]{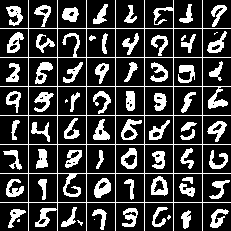}\\
(c) hard thresholding
\end{minipage}
\hfill
\begin{minipage}{.49\linewidth}
\centering
\includegraphics[width=0.75\linewidth]{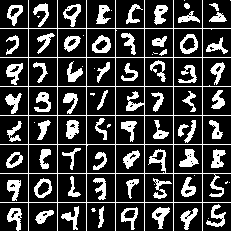}\\
(b) proposed (with SBNs)\\[1ex]
\includegraphics[width=0.75\linewidth]{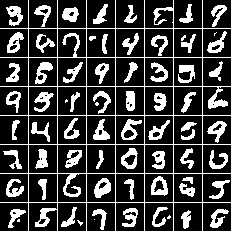}\\
(d) Bernoulli sampling
\end{minipage}
\caption{Sample generated digits for the proposed and the real-valued models, both implemented by CNNs and trained with the WGAN-GP objective. (a) and (b) show the results for the proposed model with DBNs and SBNs, respectively. (c) and (d) show the binarized results of applying a threshold of $0.5$ and Bernoulli sampling, respectively, to the raw predictions of the real-valued model.}
\label{fig:cnn}
\end{figure}

We present in \figref{fig:cnn} some sample generated digits by the proposed and the real-valued model implemented by CNNs. It can be clearly seen that the CNN model can better capture the characteristics of different digits and generate less artifacts even with a smaller number of trainable parameters as compared to the MLP model.

\section{Discussions and Conclusions}

We have presented a novel GAN-based model that can generate binary-valued predictions without any further post-processing and can be trained by end-to-end backpropagation. We have implemented such a model to generate binarized MNIST digits and experimentally compared the proposed model for different types of binary neurons, GAN objectives and network architectures. Although the results are still preliminary, we have shown that the use of gradient estimators can be a promising direction for modeling discrete distributions with GANs. A future direction is to examine the use of gradient estimators for training a GAN that has a conditional computation graph~\cite{bengio,hmrnn}, which allows the system to make binary decisions by binary neurons for more advanced designs.


\bibliographystyle{plain}
\bibliography{ref}

\begin{figure}
\centering
\setlength{\tabcolsep}{1pt}
\begin{tabular}{@{}c@{\hspace{4pt}}cccc@{}}
&DBNs &DBNs (preactivated) &SBNs &SBNs (preactivated)\\[3pt]
\raisebox{33pt}{\shortstack{WGAN-GP\\(w/ BN in $D$)}} &\includegraphics[scale=0.35]{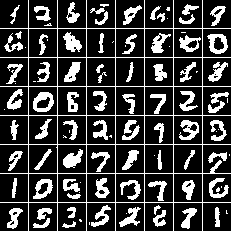} &\includegraphics[scale=0.35]{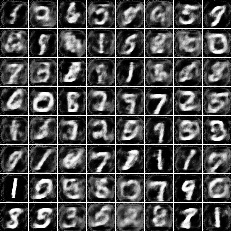} &\includegraphics[scale=0.35]{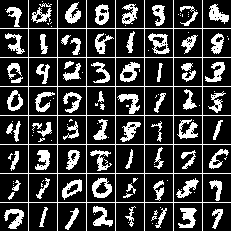} &\includegraphics[scale=0.35]{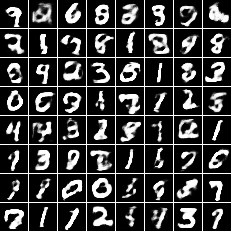}\\
\raisebox{33pt}{\shortstack{WGAN-GP\\(w/o BN in $D$)}} &\includegraphics[scale=0.35]{mlp_dbn.png} &\includegraphics[scale=0.35]{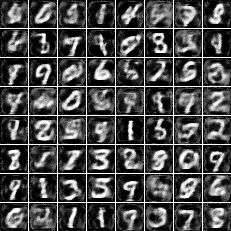} &\includegraphics[scale=0.35]{mlp_sbn.png} &\includegraphics[scale=0.35]{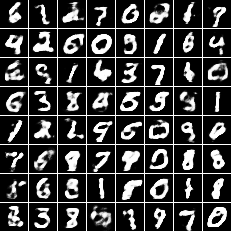}\\
\raisebox{33pt}{\shortstack{WGAN\\(w/ BN in $D$)}} &\includegraphics[scale=0.35]{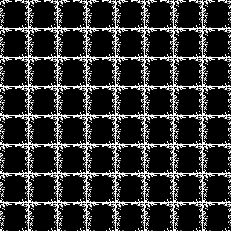} &\includegraphics[scale=0.35]{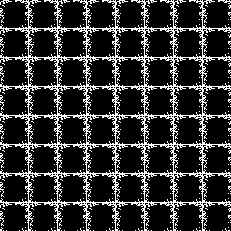} &\includegraphics[scale=0.35]{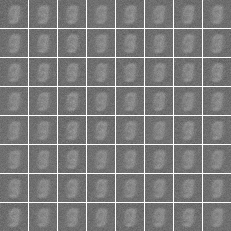} &\includegraphics[scale=0.35]{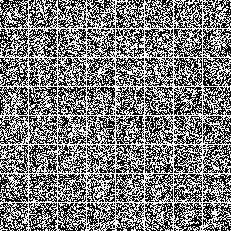}\\
\raisebox{33pt}{\shortstack{WGAN\\(w/o BN in $D$)}} &\includegraphics[scale=0.35]{mlp_dbn_wgan_no_bn.png} &\includegraphics[scale=0.35]{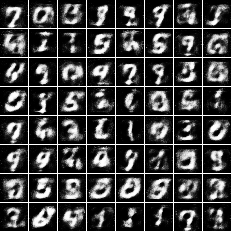} &\includegraphics[scale=0.35]{mlp_sbn_wgan_no_bn.png} &\includegraphics[scale=0.35]{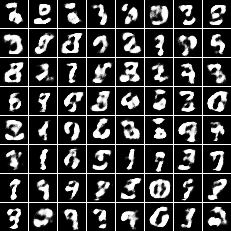}\\
\raisebox{33pt}{\shortstack{GAN\\(w/ BN in $D$)}} &\includegraphics[scale=0.35]{mlp_dbn_gan.png} &\includegraphics[scale=0.35]{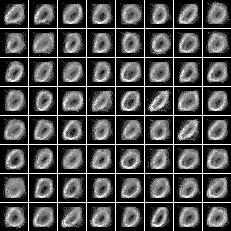} &\includegraphics[scale=0.35]{mlp_sbn_gan.png} &\includegraphics[scale=0.35]{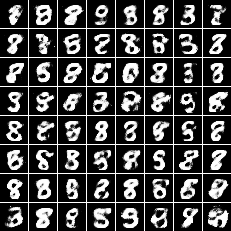}\\
\raisebox{33pt}{\shortstack{GAN\\(w/o BN in $D$)}} &\includegraphics[scale=0.35]{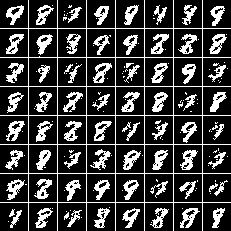} &\includegraphics[scale=0.35]{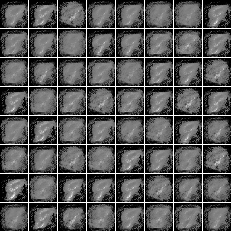} &\includegraphics[scale=0.35]{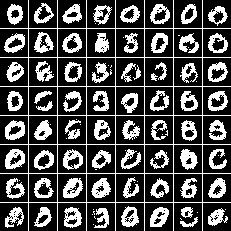} &\includegraphics[scale=0.35]{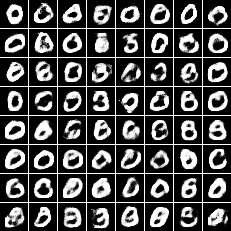}\\
\end{tabular}
\caption{Sample generated digits and preactivated outputs (i.e., the real-valued, intermediate values right before the binarization operation; see \secref{sec:bn}) for the proposed model trained with different GAN objectives. Note that `BN' stands for batch normalization.}
\label{fig:big}
\end{figure}

\end{document}